\documentclass[letterpaper, 10 pt, conference]{ieeeconf}  % Comment this line out if you need a4paper
\usepackage{cite}
\usepackage{graphicx}
\usepackage[dvipsnames]{xcolor}
\usepackage{hhline}
\usepackage{diagbox}
\usepackage{printlen}
\usepackage{amssymb}
\usepackage{wasysym} % for the diameter command 
\usepackage{units}
\usepackage{textcomp}

\usepackage{svg}
\usepackage{comment}

\usepackage{booktabs}
\newcommand{\ra}[1]{\renewcommand{\arraystretch}{#1}}

\IEEEoverridecommandlockouts                              % This command is only needed if 
\begin{document}
\title{\LARGE \bf
 An Experimental Characterization of Mechanical Layer Jamming Systems
}

\begin{comment}
\author{Authors redacted}
\end{comment}

\author{
    Jessica Gumowski$^{1, 2, 3}$, Krishna Manaswi Digumarti$^{1, 2}$ and David Howard$^{3}$%
    \thanks{J.G., and K.M.D. acknowledge continued support from the Queensland University of Technology (QUT) through the Centre for Robotics and CSIRO’s Alberto Elfes Memorial Scholarship Fund.}% <-this % stops a space
    \thanks{$^{1}$Centre for Robotics, $^{2}$School of Electrical Engineering and Robotics, Queensland University of Technology,  Brisane, 4000 QLD, Australia.}
            %{\tt\small : jessica.gumowski@qut.edu.au}}%
    \thanks{ $^{3}$CSIRO Robotics, Commonwealth Scientific and Industrial Research Organisation (CSIRO), Pullenvale, QLD 4069, Australia.}
    \thanks{Correspondence e-mail: jessica.gumowski@qut.edu.au}
            %{\tt\small b.d.researcher@ieee.org}}%
}% <-this % stops a space

\maketitle
\thispagestyle{empty}
\pagestyle{empty}

%%%%%%%%%%%%%%%%%%%%%%%%%%%%%%%%%%%%%%%%%%%%%%%%%%%%%%%%%%%%%%%%%%%%%%%%%%%%%%%%
\begin{abstract}
Organisms in nature, such as Cephalopods and Pachyderms, exploit stiffness modulation to achieve amazing dexterity in the control of their appendages.  In this paper, we explore the phenomenon of layer jamming, which is a popular stiffness modulation mechanism that provides an equivalent capability for soft robots.  More specifically, we focus on \textit{mechanical layer jamming}, which we realise through two-layer multi material structure with tooth-like protrusions. We identify key design parameters for mechanical layer jamming systems, including the ability to modulate stiffness, and perform a variety of comprehensive tests placing the specimens under bending and torsional loads to understand the influence of our selected design parameters (mainly tooth geometry) on the performance of the jammed structures. We note the ability of these structures to produce a peak change in stiffness of 5 times in bending and 3.2 times in torsion. We also measure the force required to separate the two jammed layers, an often ignored parameter in the study of jamming-induced stiffness change. This study aims to shed light on the principled design of mechanical layer jammed systems and guide researchers in the selection of appropriate designs for their specific application domains.

\end{abstract}

%%%%%%%%%%%%%%%%%%%%%%%%%%%%%%%%%%%%%%%%%%%%%%%%%%%%%%%%%%%%%%%%%%%%%%%%%%%%%%%%
\section{INTRODUCTION}

Cephalopods, including octopuses and squids, have the amazing ability to control the stiffness of their arms. This allows them to swim, walk, catch prey, squeeze through tight spaces, and even use tools such as coconut shells \cite{coconut_octopus}. Octopuses have been observed exploiting stiffness modulation to selectively stiffen sections of their arms to create temporary skeleton-like structures, subsequently allowing them to employ simplified 3DoF joint-like movements \cite{octopus_human_like} instead of having to fully control what would otherwise be a complex continuum structure. Similarly, squids and elephants use stiffness tuning in their tentacles and trunks to selectively exert strong forces at the required point \cite{Topology-architect-Oct}. Softness in the mechanical structure of the body enables safe and agile interaction with the environment, while the ability to selectively stiffen the body helps apply more force when needed. This is a key structure that facilitates the embodied intelligence of these creatures, and therefore the ability to tune stiffness is widely studied in soft robotics as a means of reaching heightened levels of embodied intelligence\cite{manti2016stiffening, yang2018principles}.

\begin{figure}[h]
  \centering
  \includegraphics[width=0.8\linewidth]{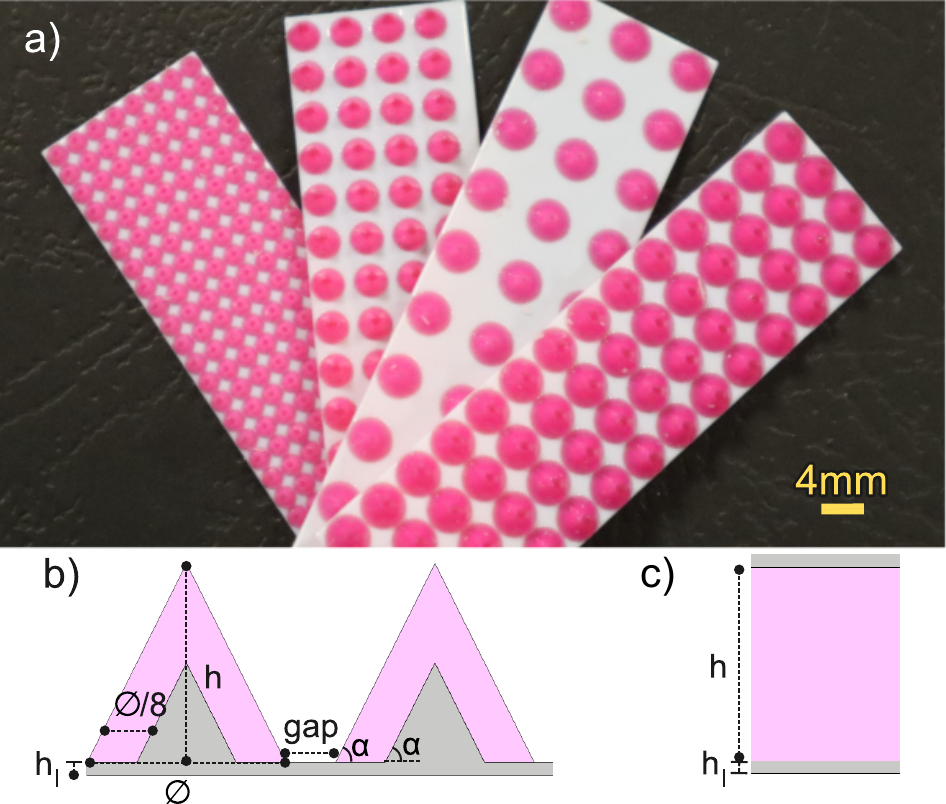}
  \caption{Design of the layer jamming structure. (a) Showing the 3D printed structure with the teeth that enable mechanical layer jamming. (ab) Cross-sectional view through the teeth showing one layer and the geometric parameters. The gap is uniform along both the length and width axes. (bc) A composite layer without any teeth. The height is the only tunable parameter.}
  \label{fig:CAD}
\end{figure}

Various strategies for modulating stiffness have been explored in the context of soft robotics, including methods based on glass transition of low melting point alloys, materials that react to electric and magnetic fields and shape memory alloys: see Manti et al. \cite{manti2016stiffening} for a comprehensive survey. A highly promising approach is through jamming \cite{fitzgerald2020review,xavier2022soft}, which is the transition of a deformable structure to one of higher rigidity due to short-range interactions between proximal constituent elements within the structure \cite{jaeger2015celebrating}.  Practically, this is achieved by forcing various structural elements within the jamming system against each other under negative pressure using a vacuum pump.  Jamming-based mechanisms are attractive because they offer a simple, fast and power efficient way to achieve a large stiffness range without requiring complex control \cite{aktacs2021modeling}. 

Grains, fibres and layers are the three main structural elements.  The specific type of structural element to use is a design choice influenced by myriad factors, including form factor, required stiffness variation, and weight constraints.  Granular jamming is perhaps the most popular \cite{brown2010universal}.  Bespoke granular jamming - the design of grain geometry and materials to achieve specific performance properties, has been explored to a great extent both \textit{in silico} and experimentally \cite{10.1145/3520304.3529072,howard2023comprehensive}. However, granular jamming is only effective against compressive loads. To design robot structures that withstand a greater variety of mechanical loads like bending and twisting, as seen in natural organisms, fibres and layers may be more suited \cite{aktacs2021modeling}. While fibres offer the most freedom to alter stiffness in 3D, they are not straightforward to manufacture (although we note recent one-shot 3D printed designs for both grains \cite{howard2022one} and fibres \cite{liow2024compliant}). 

Studies on the design of layer jammed systems are relatively few in number, and furthermore the design of bespoke layer jamming systems is particularly promising given the guarantee of a reasonable amount of surface area between the layers in which to create performance-altering functional elements.  Hence, we focus on layer jamming. Work on layer jamming has been limited to analyzing simple constituent elements, however studies have shown that exploring new geometries such as corrugations \cite{wang2023folding} and rolls \cite{sun2022electro}, leads to interesting and useful mechanical responses when jammed. Adding structured cuts to flat sheets \cite{baines2023kirigami} has also been shown to improve stiffness change. Interleaving stretchable and inextensible materials within a composite structure \cite{hexagoneJamming}, and incorporating bio-inspired scale-like elements \cite{snakeJamming} are other strategies to alter the mechanical behavior of jammed sheets.

In this paper, we define and explore the design space of layers with teeth-like protrusions as elements in a jamming structure (Fig. \ref{fig:CAD}), which is called mechanical jamming.  Related works include designs with transverse beam-like structures that span the entire width of a flat sheet have been explored in the literature to improve resistance to shear \cite{gloveJamming, clutchingStripsExoskel}. In the same vein, teeth shaped like half-spheres were also used to enhance shear resistance \cite{clutchingJammingBalls}. 

Tests on mechanical layer jamming seen in the literature are typically restricted to bending tests only.
%\textcolor{Green}{
In practice, robot arms manipulating a load are also subjected to torsion due to uneven load distribution, making this mode important to study. Our design of layers introduces compliance at the interface between layers, and as a result adhesive and frictional effect. Hence. quantifying the force required to separate layers is also of interest.%}
The focus of our paper is therefore on comprehensive evaluation, including torsional and separation testing and on a rigorous study on the influence of our chosen design parameters on the performance of mechanical layer jamming systems across those tests.

This study examines how the geometry of interlocking, dual-material teeth affects stiffness modulation. To this end, we present 

\begin{enumerate}
    \item a novel design of jamming layers incorporating teeth-shaped features to enhance inter-layer interaction,
    \item a systematic experimental setup that enables controlled testing under different loading conditions, 
    \item a comprehensive characterization of the structure’s response to bending and twisting, and
    \item a characterization of the force required to separate jammed layers. 
\end{enumerate}

We intend this study to be used by researchers and system designers to both (i) more fully understand the potential performance of a range of mechanical jammed systems and (ii) selecting an appropriate system for their specific application area.

\section{METHODS}
\subsection{Design of Layers with Teeth}
This study investigates how the geometric parameters of the teeth influence the mechanical behavior in both the unjammed and jammed states. The layer is composed of two materials: Vero (E-modulus: 2000–3000~MPa) and Agilus (Shore A: 30, elongation at break: 218\%), both printed using a Stratasys J850 printer. The substrate and the bottom of each tooth (grey portion in Fig. \ref{fig:CAD}b) is printed with the stiffer material while the tops of the teeth (pink portion in Fig. \ref{fig:CAD}b) are covered in the softer and more elastic material. This upper layer is compliant and helps align the teeth when jammed. It also provides restoring force when unjamming.

A total of 27 samples were fabricated, varying three parameters: tooth height, tooth diameter, and gap between teeth (Table \ref{table:variables}, Fig.\ref{fig:CAD}).
%CORRECTIONS
%\textcolor{Green}{
The triangular profile was selected for the teeth to facilitate reliable engagement when the two layers are pressed against each other. The conical design makes the layer isotropic in both the planar directions. The height and diameter of the tooth influence the second moment of area and the distribution of soft and stiff material, and so were parameterised to study their influence on the mechanical response. The gap was varied to evaluate the trade-off between allowing the teeth greater freedom to self-adjust and maintaining the desired stiffness ratio. In addition, because Agilus is a soft material with higher adhesion than the stiffer Vero, introducing a gap facilitates disengagement.
%}

Jamming of layers can be achieved through multiple means such as applying a vacuum, inducing electrostatic attraction or applying mechanical pressure. In this work, since the focus is on analysing the effect of layer geometry and material properties, we restrict the actuation to the more commonly accessible method of using vacuum.

\begin{table}[h!]
    \centering
    \caption{Design parameters that result in the 27 samples.}
    \begin{minipage}{.48\linewidth}
        \centering
        \ra{1.3}
        \begin{tabular}{@{}ccc@{}} \toprule
            \multicolumn{3}{c}{Variables} \\ 
            \midrule
            h [mm] & \diameter [mm] & gap [-] \\ 
            \toprule
            1 & 2 & \textonehalf \diameter \\ 
            2 & 3 & \textonequarter \diameter\\ 
            3 & 4 & 0 \\
            \bottomrule
        \end{tabular}
    \end{minipage}\hfill
    \begin{minipage}{.48\linewidth}
        \centering
        \ra{1.3}
        \begin{tabular}[t]{rc}
            \toprule
            Dimensions & [mm] \\  
            \midrule
            h\textsubscript{l} & 0.2 \\ 
            width & 16 \\ 
            length & 56 \\ 
            \bottomrule
        \end{tabular}
    \end{minipage}
    \label{table:variables}
\end{table}

\subsection{Experimental Setup}

Three types of experiments were conducted to characterise the mechanical behaviour of the structures in the jammed and unjammed state. In each test, two samples of the same geometry were placed next to each other with the teeth interlocking. The sample was then loaded to determine its mechanical response.
%\textcolor{Green}{
Our experimental approach is an alternative to using a model-based approach, where approximate material models and descriptions of their interaction often lead to a sim-to-real gap.
%}

\subsubsection{Bending Test}

\begin{figure}[h]
\vspace{6pt}
  \centering
  \includegraphics[width=\linewidth]{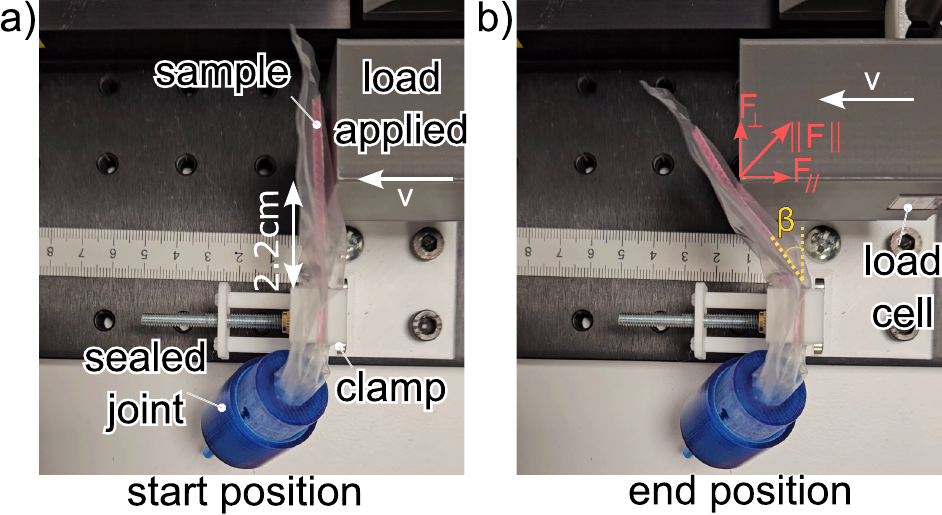}
  %\includesvg[width=\linewidth]{figures/drawing__.svg}
  \caption{Setup of the bending test (a) and the end position after a 15 mm displacement of the load (b). In this case the sample is unjammed.}
  \label{fig:bending_setup}
\end{figure}

In the first test, a bending load was applied. This test was performed in a cantilever configuration (Fig. \ref{fig:bending_setup}), where one end of the sample was clamped and the other end was subjected to a controlled load. The applied load was measured using a load cell. that recorded the resulting deformation. A 3D printed probe (made with poly-lactic acid) was mounted to the load cell to ensure a uniform pressure distribution on the sample surface. The load cell was attached to a linear stage (X-LSQ150B-E01, Zaber), which provided position feedback and controlled both displacement and velocity during the testing. The actuator applied a total displacement of 1.5~cm at a constant velocity. The data was recorded for 4 different velocities of 1, 3, 5, and 7~mm/s. Each experiment was repeated at least 3 times for statistical analysis.

The load cell measures only the force component along the direction of displacement. To determine the component of force normal to the surface of the sample, the angle of bending was determined using the position of the probe relative to the sample support and knowledge of the displacement, assuming that the change in distance between the clamp and the load cell during the testing was negligible.

The sample is enclosed in a plastic bag and sealed with a joint connected via a silicone tube to a syringe. The syringe is operated using a stepper motor to apply positive and negative pressures to the bag. This gives rise to the unjammed and jammed states respectively.

Stiffness values were computed from the force–displacement curves as interpolated slopes between 10~mm and 15~mm displacement to avoid the initial portion where the load cell interacts with the plastic bag. The slope of each curve was calculated using linear regression for each iteration, and the mean and standard deviation were computed by combining the slopes across repetitions.

\subsubsection{Twisting Test}

\begin{figure}[h]
\vspace{6pt}
  \centering
  \includegraphics[width=\linewidth]{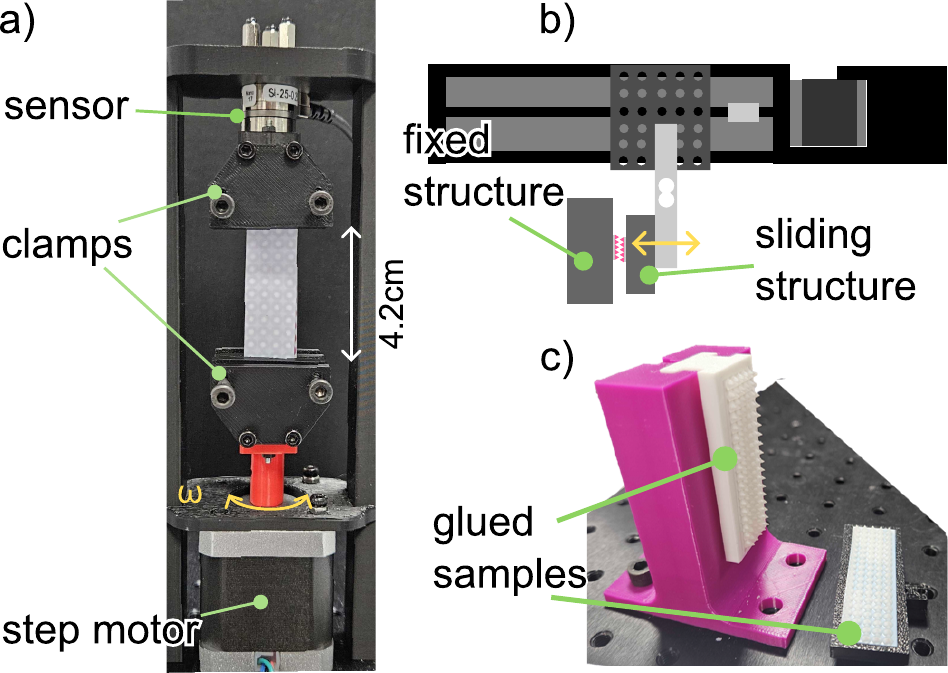}
  \caption{Setup of the twisting (a) and pulling (b) and (c) tests. The samples are glued to a support during the pulling test. The linear actuator and the load cell (b) is the same for the bending and the pulling tests.}
  \label{fig:twisting_pulling_setup}
\end{figure}

The twist test was performed using a stepper motor that applied a controlled torsional deformation, while the resulting torque was recorded using a force-torque sensor (Nano17, ATI). The sample was clamped on both ends and its alignment was adjusted using screws to minimize off-axis loading and ensure pure torsion.

During each test, the motor rotated the sample by 15$^{\circ}$ in the forward direction and then returned it to the initial position at a constant angular velocity of 1$^{\circ}$/s. The torque sensor recorded data through a data acquisition system (USB-1608G-OEM, Measurement Computing, sampling rate 1000 S/s) faster than the step rate of the motor, resulting in multiple measurements for each discrete angular position. The same setup for jamming and unjamming as used in the bending tests was used here. Each experiment was repeated at least 3 times.

\subsubsection{Separation Test}

The separation test consisted of a fixed structure on one side and a moving structure on the other. A load cell was connected to the moving side to measure the force required to separate two jammed layers. The sliding structure was actuated at a velocity of 0.5~mm/s using the same linear actuator as in the bending tests. Samples were glued to their respective printed parts using adhesive (super glue, UHU), and their alignment was carefully adjusted to minimize the angular deviation (Fig. \ref{fig:twisting_pulling_setup}). 

Each test started with a preload compressive force of 20~N mimicking the jammed stated. During testing, the actuator displaced the moving side while the load cell continuously measured the resulting pulling force. Each experiment was repeated at least 3 times. 

\section{RESULTS}
\subsection{Bending Stiffness}

\begin{figure}[h]
  \centering
  \includegraphics[width=\linewidth]{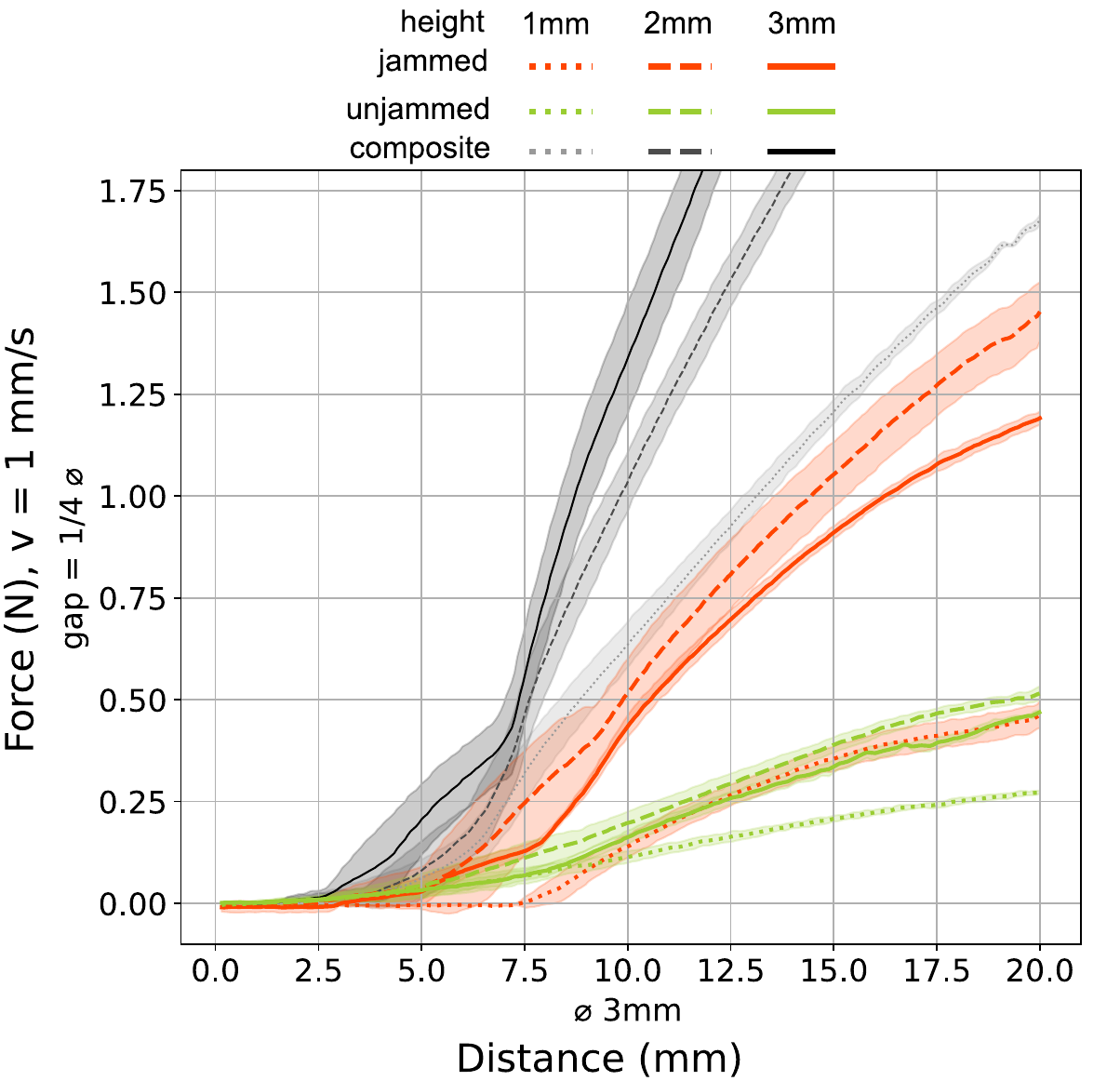}
  \caption{Force–displacement curves measured during the bending test at 1~mm/s for samples of different tooth heights. Opaque lines indicate the mean of the repetitions, and shaded areas represent one standard deviation.}
  \label{fig:bending_fig}
\end{figure}

Fig. \ref{fig:bending_fig} shows the mean force–displacement curves for the bending tests. A representative example with tooth diameter \diameter = 3 mm, and gap of 1/4\diameter for various tooth heights is shown. The shaded regions indicate one standard deviation from the mean. The recorded force was corrected for the sample angle $\beta$ (Fig. \ref{fig:bending_setup}b), computed from the measured displacement and clamp distance. The bending tests revealed that the force required to achieve a given displacement increased with the height of the teeth. Samples with taller teeth exhibited higher maximum forces, while shorter teeth deformed more easily at the same displacement.

Each repetition of the test recorded slightly different displacement values. To compute the mean and standard deviation, all data were first mapped onto a common displacement axis, ensuring alignment across iterations before statistical analysis.

\begin{figure}[h]
\vspace{6pt}
  \centering
  \includegraphics[width=\linewidth]{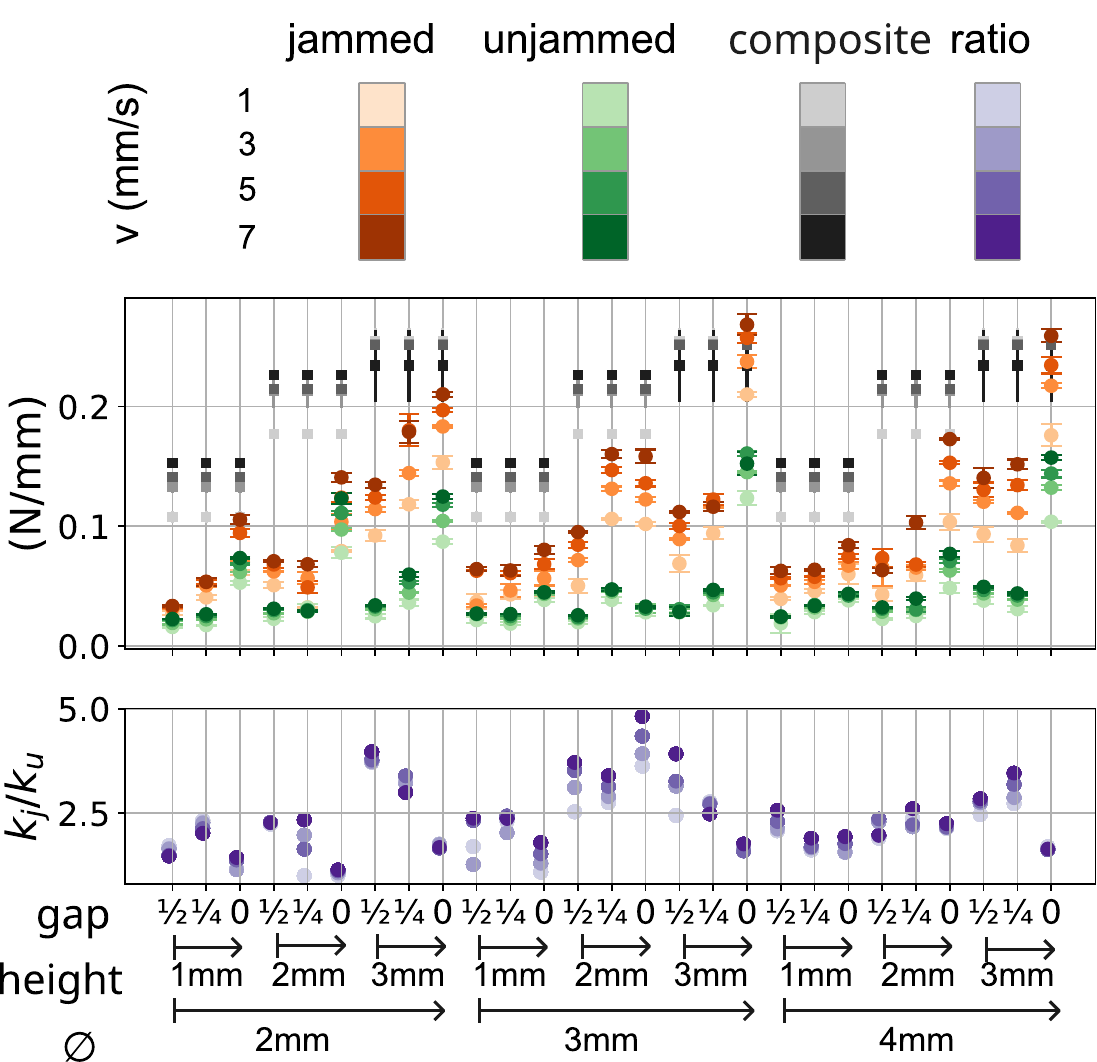}
  \caption{Stiffness values for samples measured at different velocities - for the jammed, unjammed and composite states (top) and the ratio of mean stiffness between jammed (j) and unjammed (u) states (bottom). The data were grouped by diameter to illustrate the increasing trend of stiffness with height and gap. Composite state varies only with the height and it is the same values over group diameter.}
  \label{fig:stiffness_ratio}
\end{figure}

Fig. \ref{fig:stiffness_ratio} presents the absolute bending stiffness values for both the jammed and unjammed states, as well as the ratio between them, for all samples. The results show that stiffness increased with increasing diameter and height. Jammed samples consistently show higher stiffness than unjammed ones. The data were sorted with diameter as the primary parameter, which allows observation of the trend in maximum stiffness as a function of height in the jammed state. When grouped by diameter, the stiffness increases approximately linearly with height.
The unjammed state was consistently lower than the jammed state and the jammed state was typically lower than the composite (Fig. \ref{fig:CAD}c) state. The samples with no gap between teeth showed a behaviour closest to the composite in each height class.

Fig. \ref{fig:stiffness_ratio} (bottom) shows the ratio of mean stiffness between jammed and unjammed states for each parameter variation. The jammed-to-unjammed stiffness ratio is always larger than one and peaks at 5, which indicates the contribution of the jamming state to overall sample rigidity.
In each height and diameter class, the ratio is consistently higher when the gap is $\nicefrac{1}{4}$ of the diameter.

\subsection{Torsional Stiffness}

\begin{figure}[h]
\vspace{6pt}
  \centering
  \includegraphics[width=\linewidth]{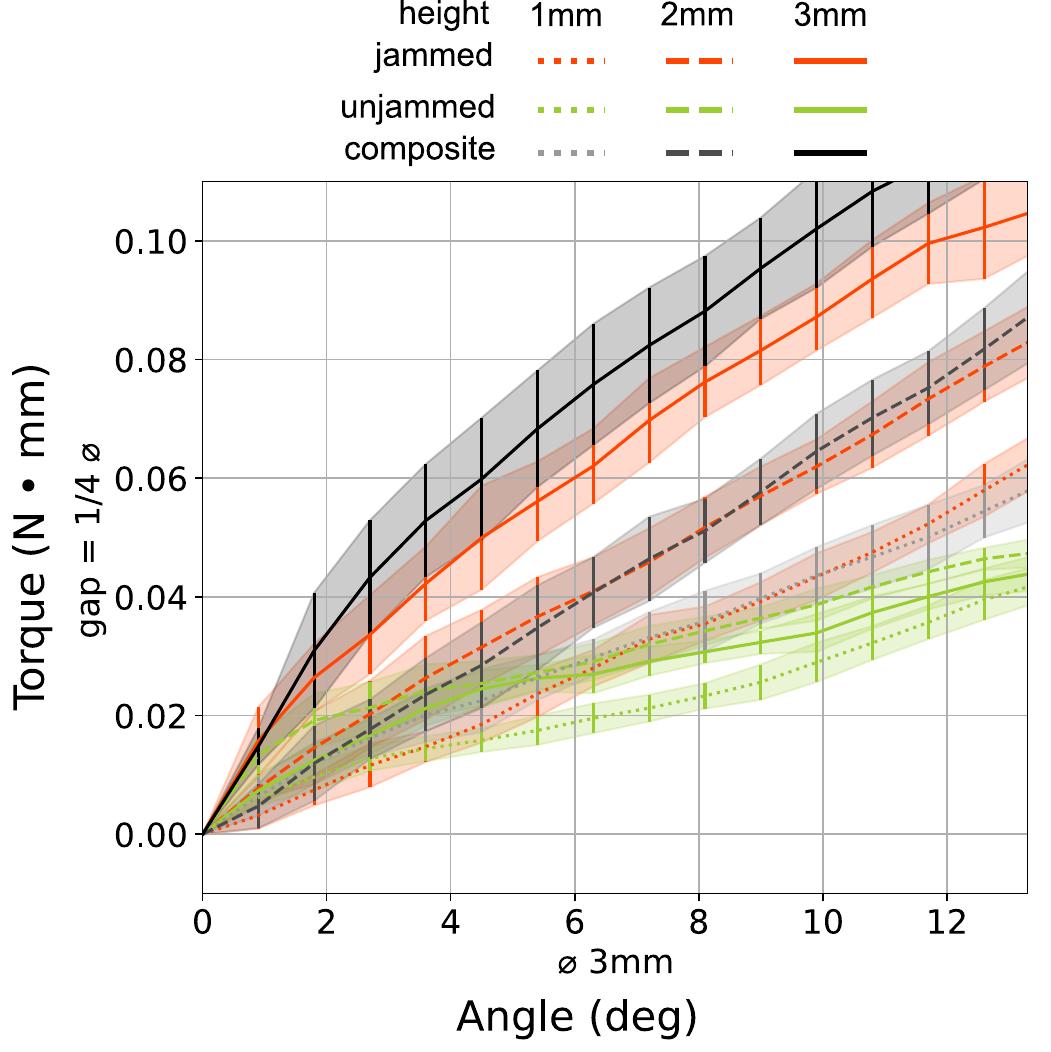}
  \caption{Torque recorded during twisting at an angular velocity of 1$^{\circ}$/s for each height variation. Each curve represents the mean torque across iterations, with error bars indicating the standard deviation at each step.}
  \label{fig:twist_raw}
\end{figure}

Fig. \ref{fig:twist_raw} shows the torque response during twisting at 1$^{\circ}$/s. A representative example with tooth diameter \diameter = 3 mm, and gap of 1/4\diameter for various tooth heights is shown. Like in the bending test, the jammed state consistently exhibited higher torque than the unjammed state. This indicates that the jamming mechanism effectively increases torsional resistance.
The jamming state approaches and sometimes overcomes the composite samples (Fig. \ref{fig:CAD}c) for equivalent height.

\begin{figure}[h]
  \centering
  \includegraphics[width=\linewidth]{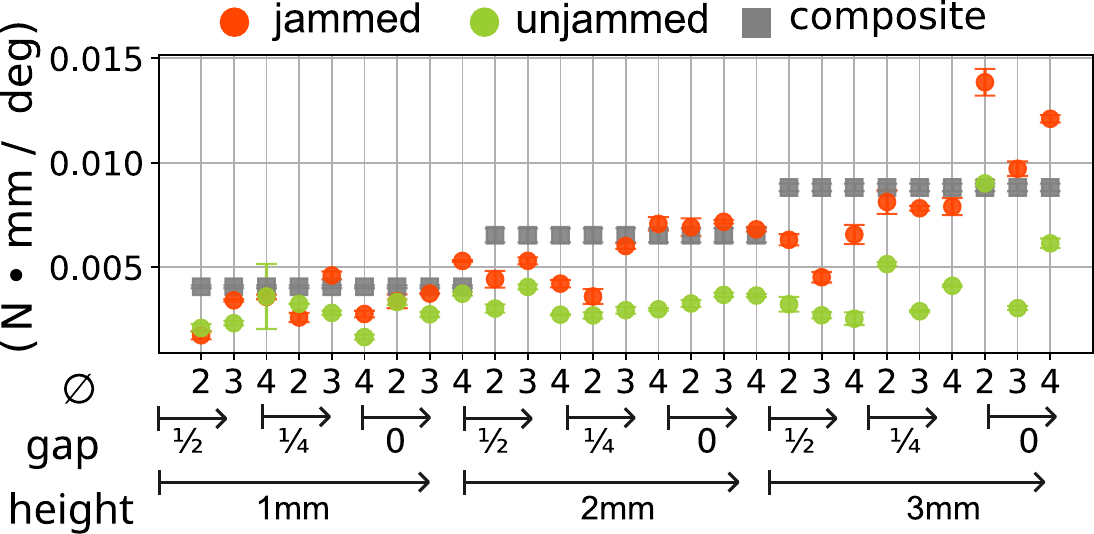}
  \caption{Comparison of the torsional stiffness of the samples. The composite sample is used as a reference to compare the jammed states. The height was chosen as the main parameter}
  \label{fig:twist_slope}
\end{figure}

The torsional stiffness, computed as the slope of the torque–angle slopes, for all the samples are compared in Fig. \ref{fig:twist_slope}. The jammed state tends to follow the behavior of the composite design, indicating a similar torsion stiffness response, while the unjammed state remains nearly constant across all parameter variations. The data is sorted with the height as the main parameter, showing a clear increasing trend of stiffness of the jammed state from the unjammed one with increasing height.

\subsection{Force of Separation}

\begin{figure}
  \centering
  \includegraphics[width=\linewidth]{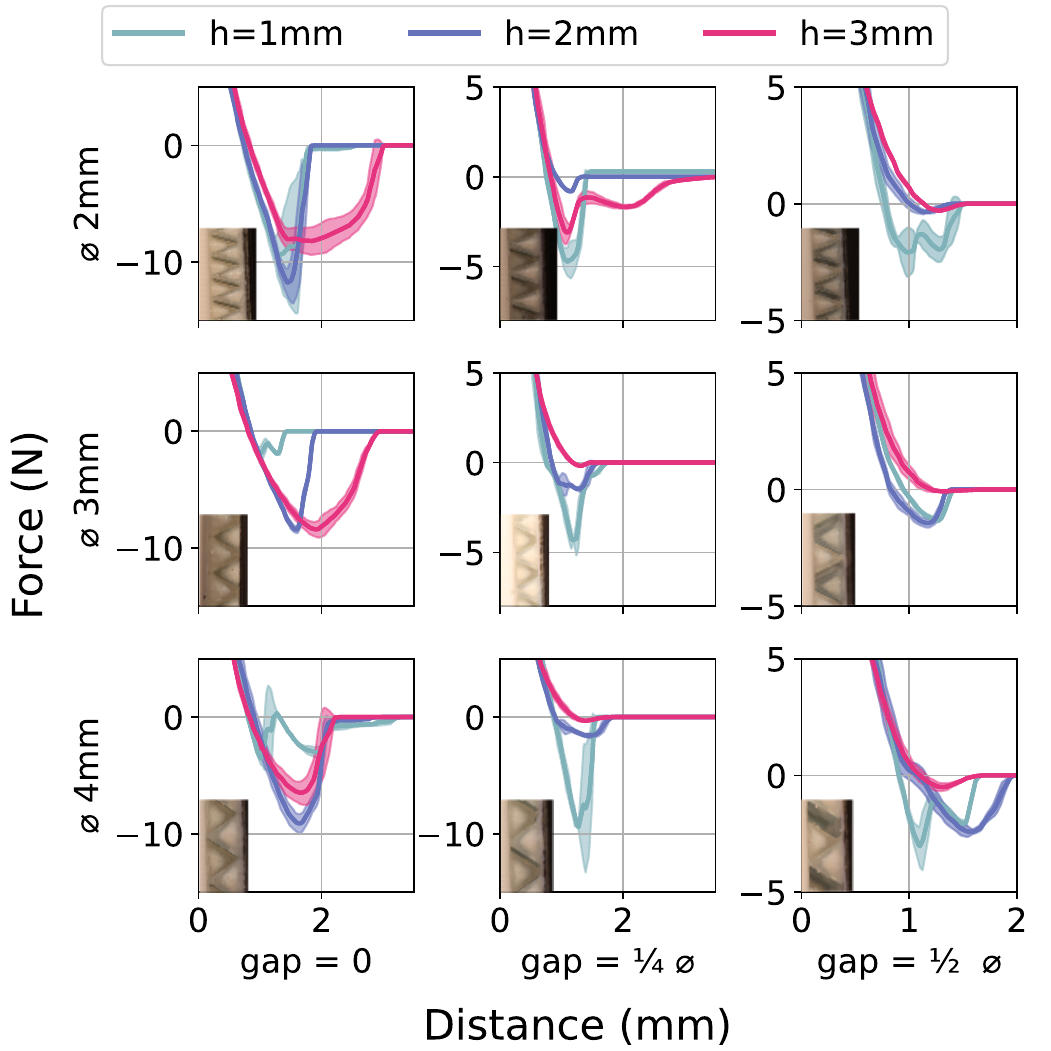}
  \caption{Force recorded during the pulling test with samples preloaded under a compression force of 20~N. The images show samples with the gap and diameter indicated on the x- and y-axes, respectively, and a height of 3~mm. Both layers are visible in the images, with the Vero component of the tooth visible through transparency and the Agilus top layer appearing translucent.}
  \label{fig:pulling_raw}
\end{figure}

Fig. \ref{fig:pulling_raw} shows the force required to fully separate two preloaded samples, as a function of displacement. The plot shows a negative peak where the force is maximum. When the curve shows several peaks, it indicates that the sample is reattaching due to misalignment. The plots indicate that the samples separate at approximately the same distance. 

These separation tests were not performed on the composite samples, as they were fabricated as a whole and a separation would mean damaging the sample.

\begin{figure}[h]
\vspace{6pt}
  \centering
  \includegraphics[width=\linewidth]{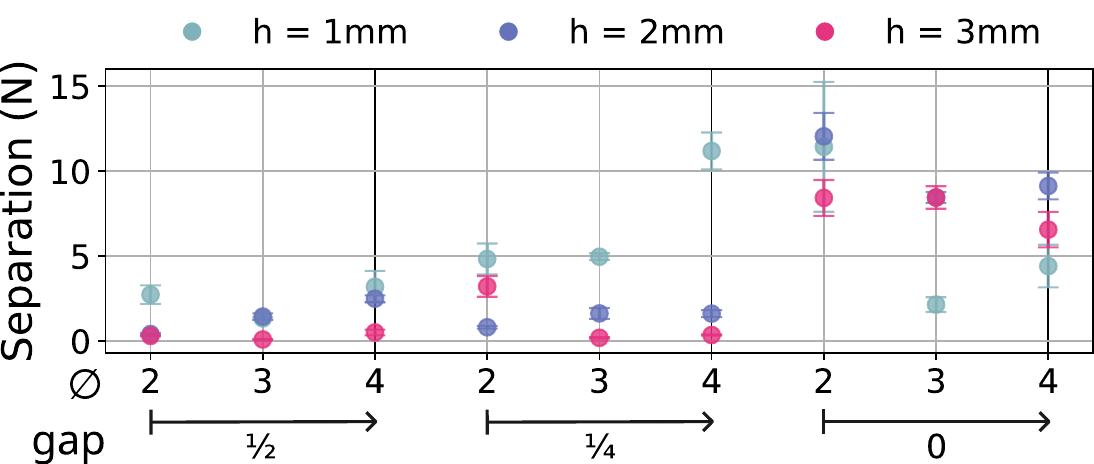}
  \caption{The maxima of the curves (Fig. \ref{fig:pulling_raw}) represent the force and relative distance, normalized by the height of the tooth, when separating two layers. In this case, the gap was selected as the primary parameter, showing that a smaller gap corresponds to a higher force.}
  \label{fig:pulling_FvsD}
\end{figure}

Fig. \ref{fig:pulling_FvsD} summarizes the maximum force recorded, with the gap as the primary parameter. The results show that the force increases as the gap decreases and as the height increases, while it is only weakly dependent on the diameter of the teeth.

\begin{table}
\vspace{12pt}
    \centering
    \caption{Angle $\alpha$ computed from the height and the diameter}
    \ra{1.3}
    \begin{tabular}{@{}cccc@{}} \toprule
        & \diameter=2~mm  & \diameter=3~mm & \diameter=4~mm \\
       \midrule
        h=1~mm & 45.0$^{\circ}$ & 33.7$^{\circ}$ & 26.6$^{\circ}$\\ 
        h=2~mm & 63.4$^{\circ}$ & 53.1$^{\circ}$ &  45.0$^{\circ}$\\ 
        h=3~mm & 71.6$^{\circ}$ & 63.4$^{\circ}$ & 56.3$^{\circ}$\\ 
        \bottomrule
    \end{tabular}
    
    \label{tab:angle_alpha}
\end{table}

\section{DISCUSSION}
When observing the change in bending stiffness, it is clear that jammed structure performs closest to the composite when the gap between teeth is zero. For smaller gaps, local shearing at the teeth due to the space available, could potentially be a reason for the reduced stiffness. As the height increases, the jammed structure is becomes stiffer, which is expected as the bending stiffness increases with increasing second moment of area.
%CORRECTIONS
%\textcolor{Green}{
Computing the moment of area would require further study and simulation due to the complexity of the geometry, which varies along the beam. In addition, the shape changes dynamically because of the material’s compliance, even though the stiffer base of the tooth (gray portion in Fig. \ref{fig:CAD}b) remains constant.
%}

%CORRECTIONS
%\textcolor{Green}{
The softer material exhibits viscoelastic effects that introduce a transient resistance before reaching the final deformation state. The effect of deformation velocity was investigated to assess how viscoelasticity influences stiffness, and the results show an increase in stiffness with increasing velocity.
%}

%CORRECTIONS
%\textcolor{Green}{
The ability to dynamically adjust stiffness offers a new perspective for soft robotic design, with potential applications across multiple domains like manipulation, grasping, locomotion and sensor design. Future work is needed to better understand and optimize this behavior.
%}

In the twisting tests, the rigid material (Vero) seems to have a higher impact on stiffness. In the jammed sample, rigid teeth are present at the interface of the two samples, while in the composite (Fig. \ref{fig:CAD}c) the intermediate material is completely soft (Agilus). This could explain the higher torsional stiffness of the jammed sample compared to the composite. The unjammed sample has a low stiffness as the layers readily separate when twisting.

The samples with shorter teeth consistently required a higher separation force. The reason remains unclear but it is speculated that the sticky nature of contact between the two soft contacting surface (Agilus on Agilus) and the angle of contact $\alpha$ could be influencing this behavior.

\section{CONCLUSIONS}
In conclusion, this study characterized the jamming of layers with dual-material teeth under two types of mechanical loading, namely bending and twisting. Three geometric parameters including the height, diameter and gap between teeth were varied to study a total of 27 parameters. Under bending load, all three parameters influenced the stiffness. A peak stiffness change of 5 times between the unjammed and jammed states was observed. Under torsion, the height of the teeth has the greatest impact on stiffness, with an observed peak of 3.2 times change in stiffness. 

We believe that our methods of characterizing specimens under various load types offer a systematic approach to evaluate and compare designs for layer jamming, including variations in geometry and material. It paves the way for future studies with multi-layered and three-dimensional arrangements of sheets in alternative configurations like cylinders and meshes within a consistent framework, enabling more interesting and useful structural designs for robots. By also focusing on the often ignored state change of unjamming, our technique to measure the force of separation informs the design of layer morphologies that enable rapid transitions between jammed and unjammed states. %\textcolor{Green}{
Other modes of loading such as shear, variation in stiffness with pressure and performance evaluation after cyclic loading are being considered in the future.
%}

\addtolength{\textheight}{-10cm}   % This command serves to balance the column lengths

%%%%%%%%%%%%%%%%%%%%%%%%%%%%%%%%%%%%%%%%%%%%%%%%%%%%%%%%%%%%%%%%%%%%%%%%%%%%%%%%

\bibliography{layerJammingv2}

\end{document}